\documentclass[letterpaper, 10 pt, conference]{ieeeconf}  % Comment this line out if you need a4paper

\IEEEoverridecommandlockouts                              % This command is only needed if 
\usepackage{microtype}
\usepackage{graphicx}
\usepackage{subfigure}
\usepackage{color}
\usepackage{xcolor}
\usepackage{amsmath} % assumes amsmath package installed
\usepackage{amssymb}  % assumes amsmath package installed
\usepackage{bigints}
\usepackage{pifont}
\def\-{\raisebox{.75pt}{-}}
\usepackage{mathtools}
\usepackage{cite}
\usepackage[makeroom]{cancel}
\usepackage{multirow}
\usepackage{hyperref}
\usepackage{booktabs}
\usepackage{subfigure}
\usepackage{epstopdf}
\usepackage{textcomp}
\usepackage{makecell}
\usepackage{bm}
\usepackage{caption}
\usepackage{algpseudocode,algorithm} % Algorithm environment
\usepackage{color, soul}
\usepackage{mathptmx}
\usepackage[11pt]{moresize}
\usepackage{hhline}

\title{\LARGE \bf
You Only Group Once: Efficient Point-Cloud Processing with Token Representation and Relation Inference Module
}

\author{Chenfeng Xu* $^{1}$, Bohan Zhai* $^{1}$, Bichen Wu* $^{2}$, Tian Li $^{3}$, Wei Zhan $^{1}$ \\ Peter Vajda $^{2}$, Kurt Keutzer $^{1}$ and Masayoshi Tomizuka $^{1}$% <-this % stops a space

\thanks{* denotes equal contribution.}% <-this % stops a space
\thanks{$^{1}$ Chenfeng Xu, Bohan Zhai, Wei Zhan, Kurt Keutzer and Masayoshi Tomizuka are with University of California, Berkeley. Correspondence to xuchenfeng@berkeley.edu}
\thanks{$^{2}$ Bichen Wu and Peter Vajda are with Facebook Reality Labs.}
\thanks{$^{3}$ Tian Li is with Peking University.}
}

\begin{document}

\maketitle
\thispagestyle{empty}
\pagestyle{empty}
\definecolor{ForestGreen}{RGB}{34,139,34}
\newcommand{\tian}[1]{\textcolor{red}{Tian:\ #1}}
\newcommand{\wbc}[1]{\textcolor{ForestGreen}{wbc:\ #1}}

%%%%%%%%%%%%%%%%%%%%%%%%%%%%%%%%%%%%%%%%%%%%%%%%%%%%%%%%%%%%%%%%%%%%%%%%%%%%%%%%
\begin{abstract}

3D point-cloud-based perception is a challenging but crucial computer vision task. A point-cloud consists of a sparse, unstructured, and unordered set of points. 
To understand a point-cloud, previous point-based methods, such as PointNet++, extract visual features through hierarchically aggregation of local features. However, such methods have several critical limitations: 1) Such methods require several sampling and grouping operations, which slow down the inference speed. 2) Such methods spend an equal amount of computation on each points in a point-cloud, though many of points are redundant. 3) Such methods aggregate local features together through downsampling, which leads to information loss and hurts the perception performance. To overcome these challenges, we propose a novel, simple, and elegant deep learning model called \textit{YOGO} (You Only Group Once). \textit{YOGO} divides a point-cloud into a small number of parts and extracts a high-dimensional token to represent points within each sub-region. Next, we use self-attention to capture token-to-token relations, and project the token features back to the point features. We formulate the mentioned series of operation as a relation inference module (RIM). Compared with previous methods, \textit{YOGO} only needs to sample and group a point-cloud once, so it is very efficient.  
Instead of operating on points, \textit{YOGO} operates on a small number of tokens, each of which summarizes the point features in a sub-region. This allows us to avoid computing on the redundant points and thus boosts efficiency.
Moreover, \textit{YOGO} preserves point-wise features by projecting token features to point features although the computation is performed on tokens. This avoids information loss and can improve point-wise perception performance. 
We conduct thorough experiments to demonstrate that \textit{YOGO} achieves at least 3.0x speedup over point-based baselines while delivering competitive classification and segmentation performance on the ShapeNetParts and S3DIS datasets.  The code is available at \url{https://github.com/chenfengxu714/YOGO.git}.

\end{abstract}

%%%%%%%%%%%%%%%%%%%%%%%%%%%%%%%%%%%%%%%%%%%%%%%%%%%%%%%%%%%%%%%%%%%%%%%%%%%%%%%%
\section{INTRODUCTION}
\begin{figure*}[!t]
    \centering
    \includegraphics[width=\textwidth]{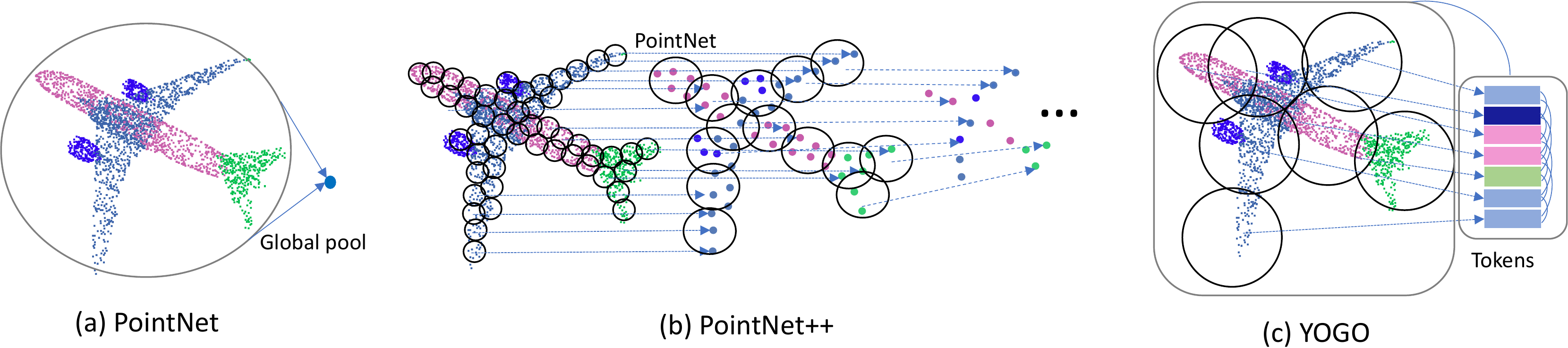}
    \caption{(a) PointNet \cite{qi2017pointnet} directly globally pools a point-cloud into a global feature. (b) PointNet++ \cite{qi2017pointnet++} hierarchically samples amounts of points, groups their nearby points, and aggregates the nearby points into one point via pointNet. (c) \textit{YOGO} divides the whole point-cloud into several parts, squeezes it into several tokens and models the token-token relations to capture the part structure of the car and the token-points relations to project the tokens into the original points.}
    \label{fig:intro}
\end{figure*}

% Lidar is a standard sensor for various applications such as automotive autopilot, augmented reality, and virtual reality. 
Increasing applications, such as autonomous driving, robotics, augmented and virtual reality, require efficient and accurate 3D perception. As LiDAR and other depth sensors become popular, 3D visual information is commonly captured, stored, and represented as point-clouds. A point-cloud is essentially a set of unstructured points. Each point in a point-cloud consists of its 3D-coordinates and, optionally, features like normal, color, and intensity. 

Different from images, where pixels are arranged as 2D grids, points in a point-cloud are unstructured. This makes it infeasible to aggregate features by using convolutions, the \textit{de facto} operator for image-based vision models. In order to handle the unstructured point-cloud, popular methods such as PointNet \cite{qi2017pointnet} process each point through multi-layer perceptrons and use a global pooling to aggregate point-wise features to form a global feature. Then, the global feature is shared with all the point-wise features. This mechanism allows points within a point-cloud to communicate local features with each other. However, such pooling mechanism is too coarse as it fails to aggregate and process local information in a hierarchical way. 
% However, it is straightforward to directly squeeze the whole point-cloud into one global feature.

Another mainstream approach, PointNet++ \cite{qi2017pointnet++}, extends PointNet \cite{qi2017pointnet} by hierarchically organizing point-cloud. Precisely, they sample center points, group nearby neighbours, and apply PointNet to aggregate local features. After PointNet++, hierarchical feature extraction \cite{li2018pointcnn,liu2019densepoint,thomas2019KPConv} has been further developed, but several limitations are not adequately addressed: 1) Sampling and grouping for point-cloud are handcraft operations, and hierarchically operating them is computationally heavy. 2) This feature extraction requires computation on all points, many of which are redundant because adjacent points don't provide extra information. 3) As local features are aggregated, point-wise features are discarded, which leads to information loss.

% Some points are more important than others, and aggregating local features for the unimportant points causes redundant computation. 2) Local neighboring features are aggregated into one point, which brings in information loss, which can be seen in Fig \ref{fig:intro} (b). 3)
% For this reason, instead of seeking to design a more advanced local-feature aggregation operator, we start from a surprising observation in the classic method PointNet \cite{DBLP:journals/corr/QiSMG16}. It max-pools the whole point-cloud into a single global feature, and combines it with the local point features, which is quite simple yet significantly enhances the performance of the MLP-based backbone. This motivates us that there may be no need to deal on each point equally since usually only the most important feature contributes to the prediction. More importantly, the recognition tasks like classification and segmentation usually depend on several local regions and their relations. For example, a car can be recognized by the front, the car windows and the wheel etc rather than scan all the car body, as shown in Fig.~\ref{fig:intro}. 

Based on the aforementioned motivation, instead of directly max-pooling the whole point-cloud into one single feature (PointNet \cite{qi2017pointnet}) or traversing to aggregating the neighbors for many points (PointNet++ \cite{qi2017pointnet++}), we propose a better way to aggregate point-cloud features, as shown in Fig \ref{fig:intro}. We only need to group points into a few sub-regions once, which has less computation cost than previous point-based methods. Then each sub-region is squeezed into a token representation \cite{wu2020visual}, which concisely summarizes the point-wise features within a region and eliminates redundancy. Next, we apply self-attention \cite{vaswani2017attention} to capture the relations between each token (region). While the representations are computed on the token-level, we also use a cross-attention module to project computed token features to point-wise features, which preserves full details with minimum computational cost. The series of operations above is what we call relation inference modules (RIM), and we propose \textit{YOGO} which 
is composed of a stack of relation inference modules.

% and our proposed YOGO is composed of a stack of relation inference modules (RIMs).

% Overall, we present \textit{YOGO} (You Only Group Once) to deal with unstructured point-cloud. In detail, \textit{YOGO} is made up of a stack of relation inference module (RIM). RIM includes a locally-squeeze module, a self-attention module and a cross-attention module, which work for squeezing point-wise features into tokens, modeling token relations, and projecting token into point-wise features.

% In detail, we first divide the point-cloud into several parts, then squeeze the points in each part into a token. We model the relations between tokens via a self-attention module \cite{vaswani2017attention,devlin2018bert} in which each element of an input is interacted with each other by a coefficient matrix. Therefore the tokens can encode the relations between each sub-region, as shown in Fig. \ref{fig:intro}. (c). To project the tokens into the original point-cloud, we use a cross-attention module which is similar to self-attention yet calculates the coefficient matrix between the points and the tokens. We term the series of operation above as a Relation Inference Module (RIM), and \textit{YOGO} is made up of a stack of RIM. \wbc{Merge this with the paragraph above. They are redundant.} 

Our proposed \textit{YOGO} can efficiently and effectively handle 3D point-cloud object classification, part segmentation, and scene segmentation. For the object classification task, \textit{YOGO} receives the raw point sets with x, y, z coordinates and will output the class based on the tokens in the final RIM layer. For the segmentation task, \textit{YOGO} also leverages the original points and outputs the per-point labels based on the final point-features. We conduct experiments on ShapeNetParts and S3DIS datasets respectively, to demonstrate the efficiency and efficacy of the proposed method. Specifically, $YOGO$ outperforms and achieves 3.0x speedup over the classic PointNet++ \cite{qi2017pointnet++}. We also provide extensive arguments for how the model works to obtain such improvements in the experiment section. 
Note that we do not expect to beat all the state-of-the-art methods via superior local-feature aggregation operators since they have evolved over several years. 
Instead, we provide a new, simple, and elegant baseline from the perspective of relational inference. We do expect it can inspire the research community to explore this approach further.

Overall, the key contributions of our work are as follows:
\begin{itemize}
    \item We design a new, simple and elegant baseline termed \textit{YOGO} which efficiently and effectively processes the unordered point-cloud.
    \item We provide thorough empirical analysis on the efficacy and efficiency of our method.
    \item We illustrate the relations extracted via the proposed \textit{YOGO} and develop intuitive explanations for its performance.
\end{itemize}

\section{Related work}
\subsection{Point-Cloud Processing}
Recent point-cloud processing methods mainly include volumetric-based method, projection-based method, and point-based method. We briefly introduce them as follows.

\textbf{Volumetric-based method} is to rasterize the point-cloud into 3D grids (voxels) \cite{maturana2015voxnet,ben20183dmfv,roynard2018classification,liu2019point} thus the convolution operator can be easily applied. However, the number of voxels is usually limited to no more than $3^3 = 27$ or $5^5 = 125$ due to the constraint of GPU memory \cite{liu2019point}. 
% This is not effective nor efficient because of the information loss and the unnecessary computations since many points are grouped into one voxel, and many empty voxels are padded, aiming to structure a regular format.
Thereby many points are grouped into the same voxel, which introduces information loss. Besides, To structure a regular format, many empty voxels are added for padding, which causes redundant computation.
Using a permutohedral lattice reduces the kernel to 15 lattices \cite{su2018splatnet}, yet the numbers are still limited. Recently, sparse convolutions \cite{liu2015sparse} are proposed to apply on the non-empty voxels \cite{choy20194d,tang2020searching,zhou2020cylinder3d,yan2018second}, largely improving the efficiency of 3D convolutions and boosting the scales of voxel numbers and the models.

\textbf{Projection-base method} attempts to project the 3D point-cloud into a 2D plane and exploit 2D convolution to extract features \cite{wang2018fusing,wu2017squeezeseg,wu2018squeezesegv2,xu2020squeezesegv3,su2015multi,lawin2017deep,boulch2017unstructured}. Specifically, the bird-eye-view projection \cite{yang2018pixor,lang2019pointpillars} and the spherical projection \cite{wu2017squeezeseg,wu2018squeezesegv2,xu2020squeezesegv3,milioto2019rangenet++} make great progresses in outdoor point-cloud segmentation (\textit{e.g.} autonomous driving). However, it is hard for them to capture the point-cloud geometry structure especially when the point-cloud becomes complex, such as indoor scenes \cite{armeni2017joint} and object parsing \cite{chang2015shapenet} since the points are collected in an irregular manner. 

\textbf{Point-based method} is to directly process the point-cloud. The most classic method PointNet \cite{qi2017pointnet} consumes the points by MLP network and squeezes the final high-dimensional features into a single global feature via global pooling, which significantly improves the performance of the point-cloud recognition and semantic segmentation. Furthermore, PointNet++ \cite{qi2017pointnet++} extends it into a hierarchical form, in each layer the local features are aggregated. Many works further develop advanced local-feature aggregation operators that mimics the convolution with customised method to structure data  \cite{li2018pointcnn,hua2018pointwise,liu2019densepoint,liu2020closer,wang2017cnn,li2018so,komarichev2019cnn}. However, the unstructured format of the point-cloud dramatically hinders their efficiency since many point-based methods rely on hierarchical handcraft operations like farthest point sampling, K nearest neighbors and ball query. Although they gradually downsample the points in each layer, the sample-group operations are required to be applied in each stage, which makes it inefficient.

Our proposed \textit{YOGO} is a new pipeline regarding the point-based method, which only needs to sample-group the points once and is robust to the various sample-group operations.

\begin{figure*}[!t]
    \centering
    \includegraphics[width=\textwidth]{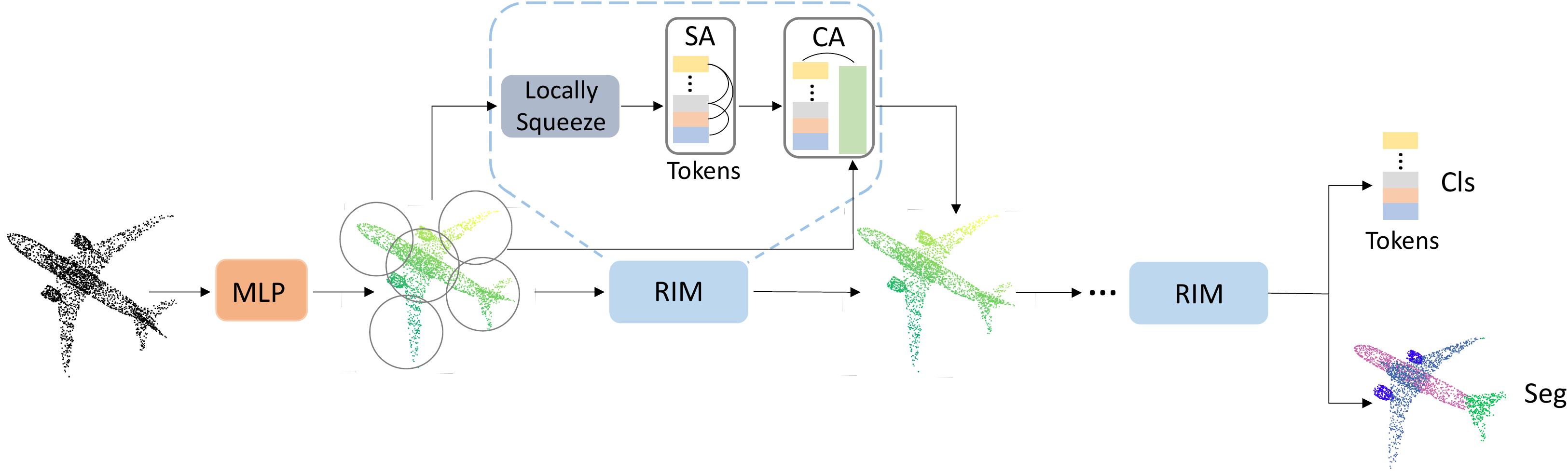}
    \caption{The framework of \textit{YOGO}. It consists of several stacked relation inference module (RIM). "Locally Squeeze" indicates that we locally squeeze the features into several tokens based on the sub-regions, "SA" means self-attention module and "CA" means cross-attention module.}
    \label{fig:framework}
\end{figure*}

\subsection{Self-attention for perception}
Self-attention was first introduced by Vaswani \textit{et al.} \cite{vaswani2017attention} for the natural language processing and recently becomes especially popular in the perception field \cite{carion2020end,dosovitskiy2020image,wu2020visual,peize2020sparse,zhu2019asymmetric}. It has great advantages to model the relations among per-elements such as tokens (for sequence data \cite{devlin2018bert,vaswani2017attention}) and pixels (for image data \cite{carion2020end}) without complex data rearrangement. This is very useful for point-cloud data because it avoids structuring the data as previous point-based method did. Yet, few works explore self-attention on point-cloud, a main challenge is that the self-attention operation is computationally intensive with the time complexity of $O(N^2)$ and makes the model unscalable. For commonly used point-cloud dataset \cite{shilane2004princeton,chang2015shapenet,armeni2017joint}, at least 1024 points are required to recognize the point-cloud, which is a relatively large number for self-attention. Thus, this motivates us to explore how to efficiently take advantage of the self-attention mechanism.

In this paper, different from the previous point-based methods that aim to design a more advanced local-feature aggregation operator, we propose a new baseline model termed \textit{YOGO} that takes advantage of the pooling operation and the self-attention. In particular, instead of directly applying the self-attention to process the point-cloud, {we use pooling operation to aggregate the features (in a sub-region) so that we only preserve important features} and can apply the self-attention in an efficient manner to model the relations between the important features. The details will be introduced in the method section.

% Attention based methods especially transformer \cite{DBLP:journals/corr/VaswaniSPUJGKP17} based on global information aggregation and low memory cost, replacing RNNs in many problems in natural language processing, and speech processing. Also in computer vision there is similar idea, called Non-local \cite{DBLP:journals/corr/abs-1711-07971} which scan through all positions in the input feature maps and find relation ship across whole sequence. Recently lots of works tried to adopt transformers in vision models , Dosovitskiy et al. propose a Vision Transformer \cite{dosovitskiy2020image}, dividing an image into 16x16 patches and feeding these patches into a standard transformer and got significant result in image classification tasks, and \cite{carion2020endtoend} also did exploration on image recognition and segmentation task, and transformer also show big capability in computer vision tasks. 

\section{Method}
% Add Task description, points segmentation + classification, define the task: "Given what kind of data  pi = (xi, yi, zi)" dimension of input data 
\subsection{Overview}
We propose \textit{YOGO}, a deep learning model for point-cloud classification and segmentation. \textit{YOGO} accepts a set of points as input $\{\mathbf{p}_i\}$, each point $\mathbf{p}_i = (x_i, y_i, z_i)$ is parameterized by its 3D coordinates and optionally, additional features such as normal, color, and so on. At the output, for classification, \textit{YOGO} generates a global prediction $\mathbf{y}$, and for semantic segmentation, it generates point-wise labels $\{\mathbf{y}_i\}$. 

\textit{YOGO} consists of a stack of relation inference modules (RIM), which is an efficient module to process a point-cloud. For a point-cloud as input $\{\mathbf{p}_i\}$ with $N$ points, we first divide it to $L$ sub-regions by uniform sampling $L$ centers and grouping their neighboring points. For points within each sub-region, we use pooling operations to extract a few tokens representing this region. The tokens are then processed by self-attention to capture the feature interactions within each region. Next, tokens are processed by a cross-attention module to exchange information among regions and aggregate long-range features. After this, we project the token features back to point-wise features, to obtain a stronger point-wise representation. 
 
% we first divide a point-cloud to several parts and feed the points into a series of RIM, as shown in Fig.~\ref{fig:framework}. In particular, we divide the point-cloud with $N$ points into $L$ sub-regions, then the point features in different sub-regions are squeezed into several tokens via the symmetric function. After this, the tokens are interacted with each other via the self-attention module in order to capture the structure of the point-cloud. Finally, the tokens will be projected into the original point features via the cross-attention modules. It is also noteworthy to mention that the tokens in prior RIM are also delivered into the current RIM, making full use of the prior information. The detail of \textit{YOGO} is illustrated as follows.
% Bohan Comments: metion L is a hyper param, and what we set for L, why L is a fix number is fine.

\begin{figure}[!t]
    \centering
    \includegraphics[width=0.5\textwidth]{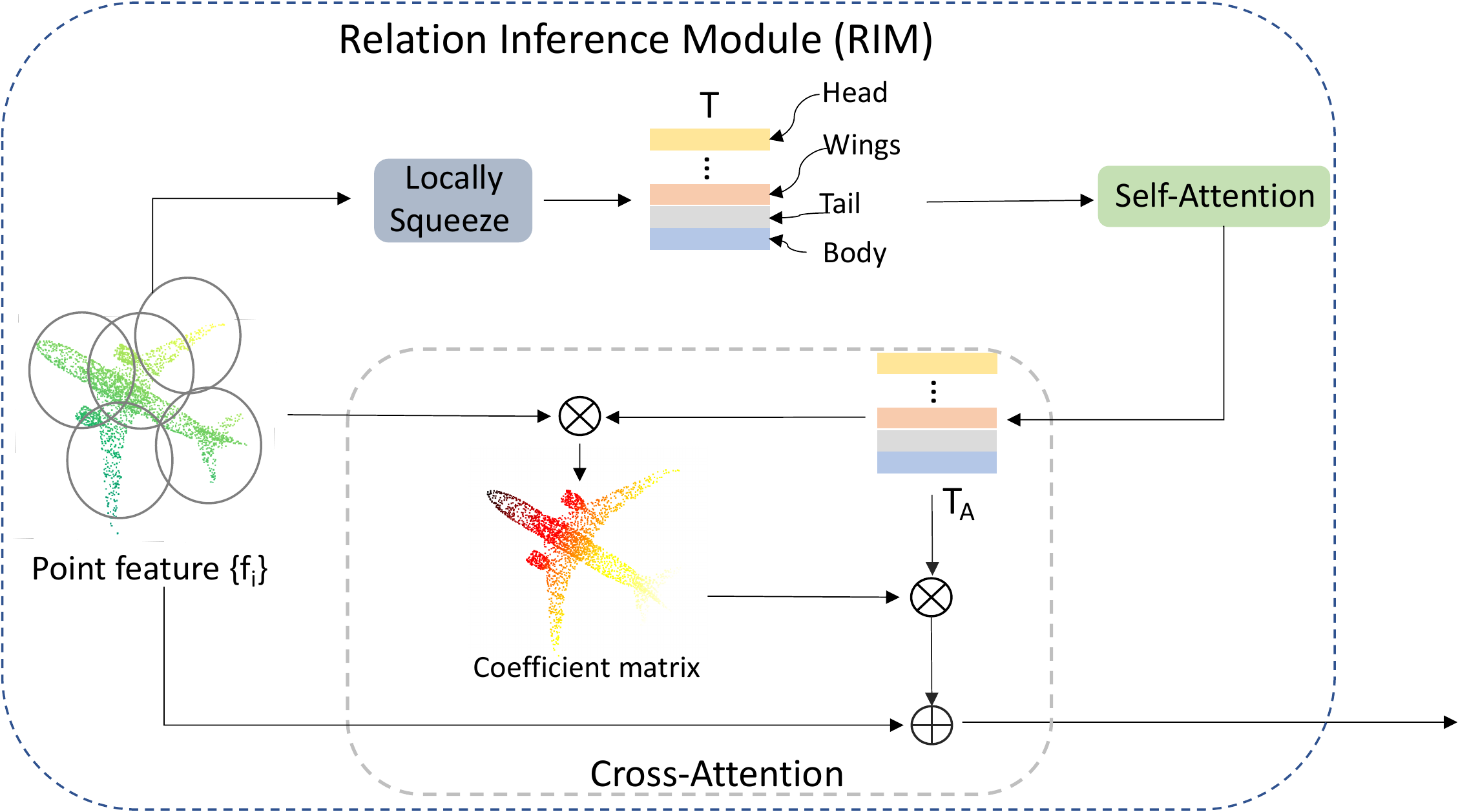}
    \caption{Relation Inference Module (RIM). The input point features $\{\mathbf{f_i}\}$ are first locally squeezed into several tokens, then we can get a stronger token representation $\mathbf{T_A}$ via self-attention module. Next, the tokens $\mathbf{T_A}$ are projected into the original point-wise features in the cross-attention module. The coefficient matrix shown in the figure is a sample from a token. }
    \label{fig:rim}
\end{figure}

\subsection{Sub-region Division}
We divide the point-cloud into $L$ ($L \ll N$) regions via the commonly used farthest point sampling (FPS) and $K$ nearest neighbouring search (KNN search) or ball query \cite{qi2017pointnet++}. Different from the previous point-based methods \cite{qi2017pointnet++,li2018pointcnn} that re-sample and re-group the points in each layer, we only need to sample-group once because the grouped indices can be re-used for gathering nearby points in each layer. Also note that the point sampling is fast since $L \ll N$ and $L$ is same in each layer. After this step, the point-cloud $\{\mathbf{p}_i\}$ is divided into $L$ sub-regions $\{\mathbf{p}_i^{(j)}\}$ where $\mathbf{p}_i^{(j)}$ denotes the $i^{th}$ point in the $j^{th}$ sub-region.

\subsection{Relation Inference Module (RIM)}
The relation inference module (RIM) does not require handcrafted grouping and sampling operations to aggregate point-wise features. Instead, it adopts a simple pooing operation an self-attention. The pooling operation is used to aggregate features within a sub-region \cite{qi2017pointnet,liu2020closer}, while self-attention is used to capture relations between sub-regions. Self-attention has been demonstrated to have a superior ability to capture long-range relations in many vision and language tasks. However, as self-attention's computational cost is quadratic with number of input elements, it is infeasible to be directly used for point-cloud. In our work, our design is to let self-attention operate on region/tokens, instead of points, this greatly reduces the computational cost while leveraging the advantages of self-attention.  The structure of RIM is presented in Fig. \ref{fig:rim}.

Specifically, given a set of point features $\{\mathbf{f}_i\}$ corresponding to a point-cloud $\{\mathbf{p}_i\}$, we divide the points into sub regions  $\{\mathbf{p}_i^{(j)}\}$ with corresponding features $\{\mathbf{f}_i^{(j)}\}$. RIM computes a token to represent points within each region as
\begin{equation}
\begin{gathered}
    \mathbf{T}^{(j)} = G(\mathbf{ \text{maxpool}_{i} (\{\mathbf{f}_i^{(j)}\})}), 
\end{gathered}
\end{equation}
where $\mathbf{f}_i^{(j)} \in \mathbb{R}^{1 \times C}$ denotes point-$i$ in region-$j$. $\text{maxpool}_i$ is the max-pooling operation to squeeze points in $\{\mathbf{f}_i^{(j)}\}$ to one. $G(\cdot)$ is a linear function that maps the pooled feature to the output $\mathbf{T}^{(j)} \in \mathbb{R}^{1\times C_T}$. 
This is similar to PointNet \cite{qi2017pointnet}, yet PointNet applies a pooling to all the points, while we focus on different sub-regions.

Next, to model the relations between different regions using self-attention as \cite{vaswani2017attention}. We first combine all the tokens $\{\mathbf{T}^{(j)}\}$ to form a tensor $\mathbf{T} \in \mathbb{R}^{L\times C_T}$. Since self-attention is a permutation-invariant operation, we can choose any order to stack $\{\mathbf{T}^{(j)}\}$ to form $\mathbf{T}$. On the token, we compute
\begin{equation}
\begin{gathered}
M_{tt} = \mathbf{T}\hat{W_k}(\mathbf{T}\hat{W_q})^\mathrm{T}, \\
SA(\mathbf{T}) = softmax(M_{tt})(T\hat{W_v}) + \mathbf{T},\\
\mathbf{T_A} = SA(\mathbf{T})\hat{W_p}.
\end{gathered}
\end{equation}
where $\mathbf{T_A} \in \mathbb{R}^{L \times C_T}$ is the output of the self-attention module, $M_{tt} \in \mathbb{R}^{L \times L}$ is the coefficient matrix regarding token-token relation in $\mathbf{T} \in \mathbb{R}^{L \times C_T}$, and $\hat{W_p} \in \mathbb{R}^{C_T \times C_T}$, $\hat{W_v} \in \mathbb{R}^{C_T \times C_T}$, $\hat{W_k} \in \mathbb{R}^{C_T \times C_T}$, $\hat{W_q} \in \mathbb{R}^{C_T \times C_T}$ are the parameterized matrix, representing the weights for project, value, key and query \cite{vaswani2017attention}, respectively.

Next, we project the output tokens back to point-wise features. To do this, we feed $T_A \in \mathbb{R}^{L \times C_T}$ and the original point-wise feature $\{\mathbf{f}_i\}$ to a cross-attention module as
\begin{equation}
    \begin{gathered}
         \mathbf{\bar{T}_A} = G'(\mathbf{T_A}), \\
         M_{tp} = \mathbf{f}_iW_k(\mathbf{\bar{T}_A}W_q)^\mathrm{T},\\
         CA(\mathbf{\bar{T}_A},\mathbf{f}_i) = softmax(M_{tp})(\mathbf{\bar{T}_A}W_v) + \mathbf{f}_i, \\
         \mathbf{\bar{f}}_i = CA(\mathbf{\bar{T}_A},\mathbf{f}_i)W_p,
    \end{gathered}
\end{equation}
where $\mathbf{\bar{f}}_i$ is the output point-feature of the cross-attention module, which is parametrized by the matrix $W_p \in \mathbb{R}^{C \times C}$, the matrix $W_v \in \mathbb{R}^{C \times C}$, the matrix $W_k \in \mathbb{R}^{C \times C}$, and the matrix $W_q \in \mathbb{R}^{C \times C}$. $G'$ is a MLP mapping the $\mathbf{T_A} \in \mathbb{R}^{L \times C_T}$ into $\mathbf{\bar{T}_A} \in \mathbb{R}^{L \times C}$ and  $M_{tp} \in \mathbb{R}^{L \times 1}$ is a coefficient matrix between the tokens $\mathbf{\bar{T}_A}$ and a point feature $\mathbf{f}_i$.

Analytically, each token element in $\mathbf{T_A}$ encodes the most important and representative components in each corresponding sub-region. The self-attention regarding the $\mathbf{T_A}$ can help capture the structure of the point-cloud, \textit{e.g.}, as shown in Fig. \ref{fig:rim}, the head, the wings, the airplane body, and the tail have strong structure relations between each other. This is useful to learn semantics since we can recognize the airplane through these representative elements. After this, the coefficient matrix $M_{tp}$ is calculated to indicate the relations between the tokens and each point which helps to project tokens with rich semantic information into original point-cloud features.

\subsection{Training objective}
Our proposed network \textit{YOGO} can handle point-cloud classification and segmentation tasks and can be simply trained end-to-end. The tokens and the point feature in the final $RIM$ are used for classification and segmentation tasks, respectively. We simply apply the commonly used cross-entropy loss to train the model for both classification and segmentation task, which are respectively given by
\begin{equation}
 L_{cls}= - \sum_{c=1}^C \hat{y_c} \cdot log(p_c),
\end{equation}
where $L_{cls}$ is the cross-entropy loss for classification, $C$ is the number of categories of a point-cloud, $\hat{y_c}$ is the ground truth and $p_c$ is the softmax probability prediction to the $c^{th}$ class, then
\begin{equation}
     L_{seg} = \frac{- \sum_{N} \sum_{c=1}^C {y_n}_c \cdot log({p_n}_c)}{N},
\end{equation}
where $L_{seg}$ is for segmentation, $C$ is the total categories of each point, $N$ is the whole number of the point-cloud, ${y_n}_c)$ is the ground-truth for $n^{th}$ point, and ${p_n}_c$ is the softmax probability prediction.

\section{Experiment}
The experiments are conducted on the ShapeNetParts \cite{chang2015shapenet} and S3DIS \cite{armeni2017joint} for segmentation task. We conduct both comparison experiments for effectiveness and efficiency evaluation and ablation study for deep analysis of why \textit{YOGO} works. The details are illustrated as follows.
% \subsection{Dataset}
% %Bohan Todo: ModelNet40, ShapeNet, S3DIS.

% \textbf{Semantic Segmentation.} We perform semantic segmentation experiments on ShapeNet \cite{Yi16} and S3DIS \cite{DBLP:journals/corr/ArmeniSZS17} dataset. ShapeNet dataset involves 16681 different objects with 16 categories, each of which has 2-6 part labels. S3DIS dataset is collected from the real-word indoor scenes, including 3D scans of Matterport scanners from 6 areas. 

% \textbf{Classification.} We also evaluate our method for classification task on the ModelNet40 \cite{DBLP:journals/corr/WuSKTX14} dataset. It contains 12,311 meshed CAD models from 40 categories.

\begin{figure*}[!t]
    \centering
    \includegraphics[width=0.9\textwidth]{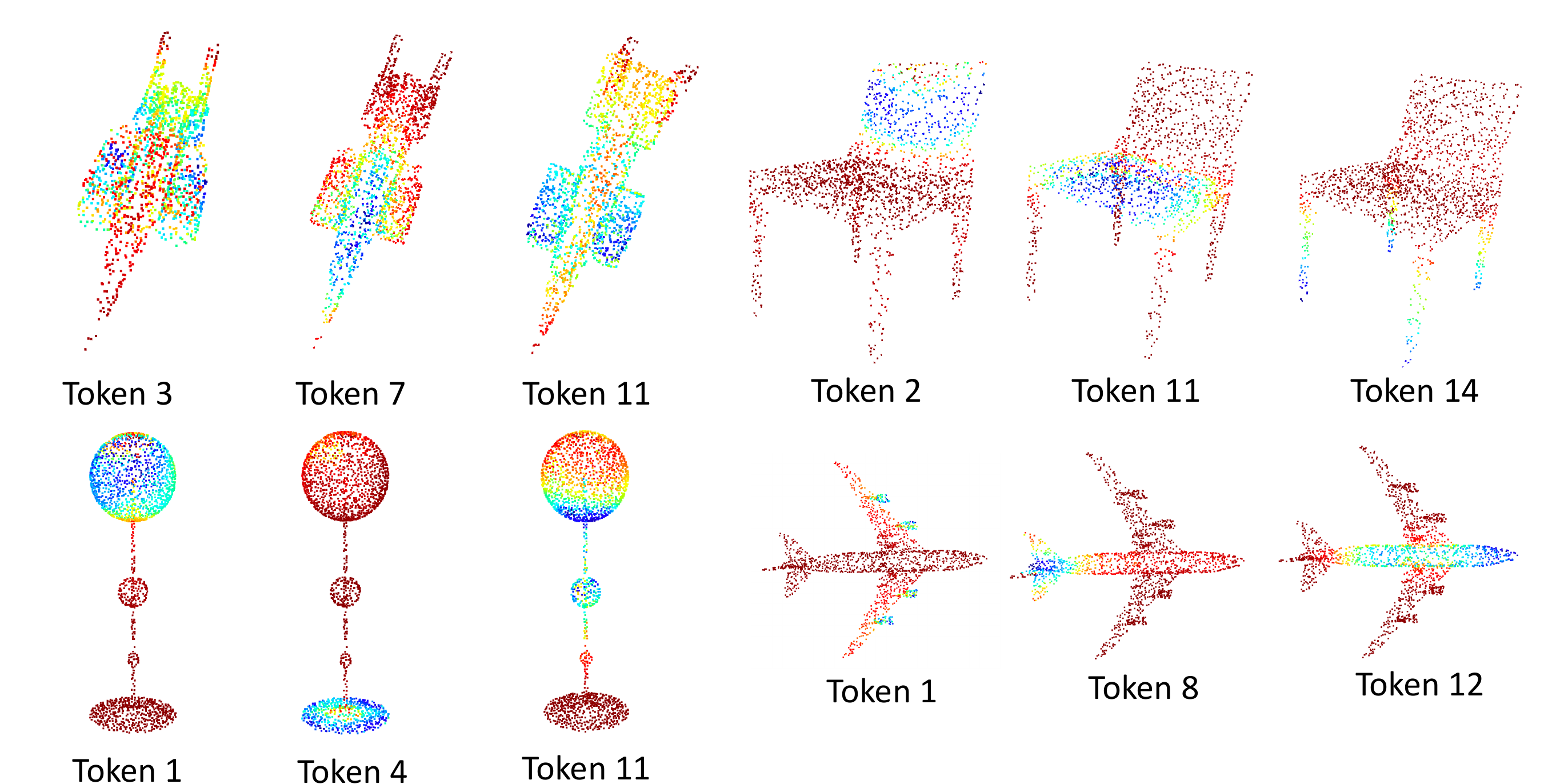}
    \caption{Visualization of the coefficient matrix in cross-attention module on the ShapeNetParts dataset. Blue means the response is small and Red means the response is large. Taking the rocket point-cloud and the chair point-cloud as examples, as for the rocket point-cloud, we choose token 3, token 7, and token 11 that attend the head, the tail, and the tail, respectively. As for the chair point-cloud, we choose token 2, token 11, and token 14 that attend the sitting board and the legs, the back board and the legs, the back board and the stting board, respectively. }
    \label{fig:vis_exp}
\end{figure*}

\subsection{Implementation detail}
We stack eight relation inference modules on all experiments and set $C_T$ equal to 256 and $C$ equal to {32, 32, 64, 64, 128, 128, 256, 256} respectively for the eight relation inference modules.

We first conduct the experiment on the ShapeNetParts \cite{chang2015shapenet} dataset, which involves 16681 different objects with 16 categories and 2-6 part labels for each. By utilizing the evaluation metric mean intersection-over-union (mIoU), we first calculate the part-averaged IoU for each of the 2874 test models and average the values as the final metrics. We set the $L$ equal to 32, $K$ equal to 96. All the coordinates of points are normalized into [0, 1]. As for the ball query grouping, we set the radium to 0.2. During training, we input the data with the size of 2048 points and the batch size of 128, and use the cosineAnnealing learning rate strategy with initial learning rate 1e-3 and Adam optimizer to optimize the network. During the inference, we input the points with the size of 2048 and vote 10 times for each point-cloud.

We then conduct the experiment on the S3DIS \cite{armeni2017joint}, which is collected from the real-world indoor scenes and includes 3D scans of Matterport Scanners from 6 areas. We also use the mIoU as the evaluation metric. We apply the 1, 2, 3, 4, 6 areas for training and use the 5 area for the test since it is the only area that does not overlap with any other area. We set the $L$ equal to 32, $K$ equal to 128. For ball query grouping, the radium is set to 0.3. We train the networks via feeding the input with the 4096 points and batch size 64, and utilize the same training strategy as the aforementioned for ShapeNetParts dataset. During the inference, we also input the points with the size of 4096 and vote one times for each point-cloud. 

Finally, we apply our method to the classification tasks on the ModelNet40 dataset \cite{shilane2004princeton}. This is a CAD dataset with 12311 meshed models and contains 40 categories. We set $L$ equal to 16, $K$ equal to 128. For ball query grouping, the radium is set to 0.15. We train the network with the input of 1024 points and batch size 64, and use the same training strategy as \cite{li2018pointcnn}. Note that all the experiments are conducted on one Titan RTX GPU. 

\subsection{Comparison experiment}
We first perform \textit{YOGO} on the ShapeNetParts and S3DIS datasets for semantic segmentation evaluation, as shown in Table. \ref{tab:shapnet} and Table. \ref{tab:s3dis}. We can observe that \textit{YOGO} has very competitive performance and speedups at least 3x over point-based baselines on the ShapeNetParts dataset. Precisely, \textit{YOGO} slightly outperforms the classic baseline PointNet++ \cite{qi2017pointnet++} and performs at least 3x faster. Although the unofficial PointNet++ \cite{yanx} slightly outperforms \textit{YOGO}, \textit{YOGO} largely speedups it over 9.2x faster. As for the S3DIS dataset, \textit{YOGO} also achieves at least 4.0x speedup and delivers competitive performance. PointCNN \cite{li2018pointcnn} outperforms \textit{YOGO} over 0.9 mIoU (\textit{resp.} 3.26 mIoU) on the ShapeNetParts (\textit{resp.} S3DIS) dataset but is not efficient. Regarding the classification task conducted on the ModelNet40 dataset presented as Table. \ref{tab:modelnet}, \textit{YOGO} still has competitive performance over the popular baselines. 

We can also observe that \textit{YOGO} with two different grouping method KNN and ball query have similar performance, which shows that our proposed method is relatively stable with regard to different grouping strategies. Note that there remain many excellent methods that perform better than \textit{YOGO}, we are not attempting to beat all of them. Instead, we propose a novel baseline from the perspective of token-token relation and token-point relation.

\begin{table}[!t]
\centering
\small
\begin{tabular}{ c c c c c }
\hline
Method  & Mean IoU & Latency & GPU Memory\\
\hline
 PointNet \cite{qi2017pointnet} & 83.7 &  21.4 ms& 1.5 GB  \\
 RSNet \cite{huang2018recurrent}  &  84.9 & 73.8 ms & 0.8 GB \\
 SynSpecCNN \cite{yi2017syncspeccnn} & 84.7 & - & - \\
 PointNet++ \cite{qi2017pointnet++}  &  85.1 & 77.7 ms & 2.0 GB \\
 PointNet++* \cite{qi2017pointnet++}  &  85.4 & 236.7 ms & 0.9 GB \\
 DGCNN \cite{wang2019dynamic}  &  85.1 & 86.7 ms& 2.4 GB\\
 SpiderCNN \cite{xu2018spidercnn} & 85.3 & 170.1 ms & 6.5 GB \\
 SPLATNet \cite{su2018splatnet} & 85.4 & -  & -   \\
 SO-Net \cite{li2018so} & 84.9 & - & - \\
 PointCNN \cite{li2018pointcnn} &  86.1 &134.2 ms & 2.5 GB \\
\hline
$YOGO$ (KNN) &  85.2 & 25.6 ms & 0.9 GB\\
$YOGO$ (Ball query) & 85.1 & 21.3 ms & 1.0 GB\\
\hline

\end{tabular}
\caption{Quantitative results of semantic segmentation on the ShapeNetPart dataset. The latency and GPU memory are measured under the batch size of 8 and the point number of 2048. PointNet++* is a reproduced version from the popular repository \cite{yanx}. }
\label{tab:shapnet}
\end{table}

\begin{table}[!t]
    \centering
    \small
    \begin{tabular}{c c  c  c c}
    \hline
    Method &  Mean IoU & Latency & GPU Memory\\
    \hline
    PointNet \cite{qi2017pointnet}& 42.97 & 24.8 ms & 1.0 GB\\
    DGCNN \cite{wang2019dynamic}  & 47.94 & 174.3 ms & 2.4 GB\\
    RSNet \cite{huang2018recurrent} & 51.93 & 111.5 ms & 1.1 GB\\
    PointNet++* \cite{qi2017pointnet++} & 50.7 & 501.5 ms & 1.6 GB\\
    TangentConv \cite{tatarchenko2018tangent} &52.6& - & - \\
    PointCNN \cite{li2018pointcnn} &  57.26 &282.43 ms & 4.6 GB \\
    \hline
    $YOGO$ (KNN) & 54.0 & 27.7 ms & 2.0 GB\\
    $YOGO$ (Ball query)& 53.8 & 24.0 ms & 2.0 GB\\
    \hline
    \end{tabular}
    \caption{Quantitative results of semantic segmentation on the S3DIS dataset. The latency and GPU memory are measured under the batch size of 8, the point number of 4096. PointNet++* is a reproduced version from the popular repository \cite{yanx}.}
    \label{tab:s3dis}
\end{table}

\begin{table}[!t]

\centering
\small
\begin{tabular}{ c c }
 \hline
 Method & Overall Accuracy \\
 \hline
 PointNet \cite{qi2017pointnet} & 89.2\\
 PointNet++ \cite{qi2017pointnet++}&  90.7\\
 SO-Net \cite{li2018so} &  90.9\\
 SpiderCNN \cite{xu2018spidercnn} &  90.5\\
 MCConv \cite{hermosilla2018monte} &  90.9 \\
 PointCNN \cite{li2018pointcnn} & 92.2 \\
 DGCNN \cite{wang2019dynamic} & 92.2 \\
 ViT-B-2* \cite{dosovitskiy2020image} & 78.9 \\
 Transformer* \cite{vaswani2017attention} & 82.1 \\
 Perceiver \cite{jaegle2021perceiver} & 85.7 \\
 \hline
\textit{YOGO} (KNN)& 91.4\\
\textit{YOGO} (Ball query)& 91.3\\
\hline 
\end{tabular}
\caption{Quantitative results on the ModelNet40 dataset. The results of ViT-B-2* and Transformer* are directly taken from Perceiver \cite{jaegle2021perceiver}.}
\label{tab:modelnet}
\end{table}

\subsection{Ablation study}
\label{ab_ov}
In this subsection, we do an in-depth study of our proposed \textit{YOGO}, including the effectiveness of the Relation Inference Module (RIM) and related techniques on the ShapeNetParts Parts dataset. More importantly, we analyze what RIM learns, thus explicitly explain why \textit{YOGO} works.

\textbf{The effectiveness of RIM.} The self-attention module and the cross-attention module are the two key components in RIM. We study them by removing them respectively. In particular, when we remove the self-attention module (SA), we insert several MLP layers that have similar FLOPs as the self-attention modules. When we remove the cross-attention module, we use different related techniques: 1) Directly squeeze the tokens $T_A$ into one global feature and concatenate it with the point feature $F_P$, which is similar to PointNet \cite{qi2017pointnet}; 2) add or concatenate the tokens $T_A$ into the $F_P$ corresponding to different sub-regions. The results are shown in Table.~\ref{tab:saca}. 

We can observe that if we coarsely use the tokens by squeezing them into one global feature, the performance is largely worse than the original \textit{YOGO} by 1.1 mIoU. A more reasonable method is to leverage them on the corresponding sub-regions by adding/concatenating each token with the corresponding point features of sub-regions. It can be seen that the original \textit{YOGO} still outperforms them by at least 0.5 mIoU, which shows the cross-attention method is more effective. On the other hand, when we substitute the self-attention module with several MLP, the performance also drops 0.5 mIoU, which also shows that it is important to learn the relations between different sub-regions.

\textbf{Visualization analysis of RIM.} To deeper explore what RIM learns, we visualize some samples of the coefficient matrix in the cross-attention module, as shown in Fig. \ref{fig:vis_exp}. Note that the coefficient matrix of the cross-attention module indicates the relation between the tokens and the original point features, yet there is no explicit supervision to guide the learning coefficient matrix. We can observe that different tokens can automatically attend to different parts even without supervision. For example, the tokens regarding the rocket point-cloud have stronger responses to the head, the tail, and the flank, respectively. Similarly, the tokens regarding the chair point-cloud have stronger responses to the back, the sitting board, and the legs, respectively. These can demonstrate that tokens indeed carry strong semantic information, and the cross-attention module indeed helps the tokens with rich semantic information be projected into point-wise features.

\begin{figure}[!t]
    \centering
    \includegraphics[width=8.5cm]{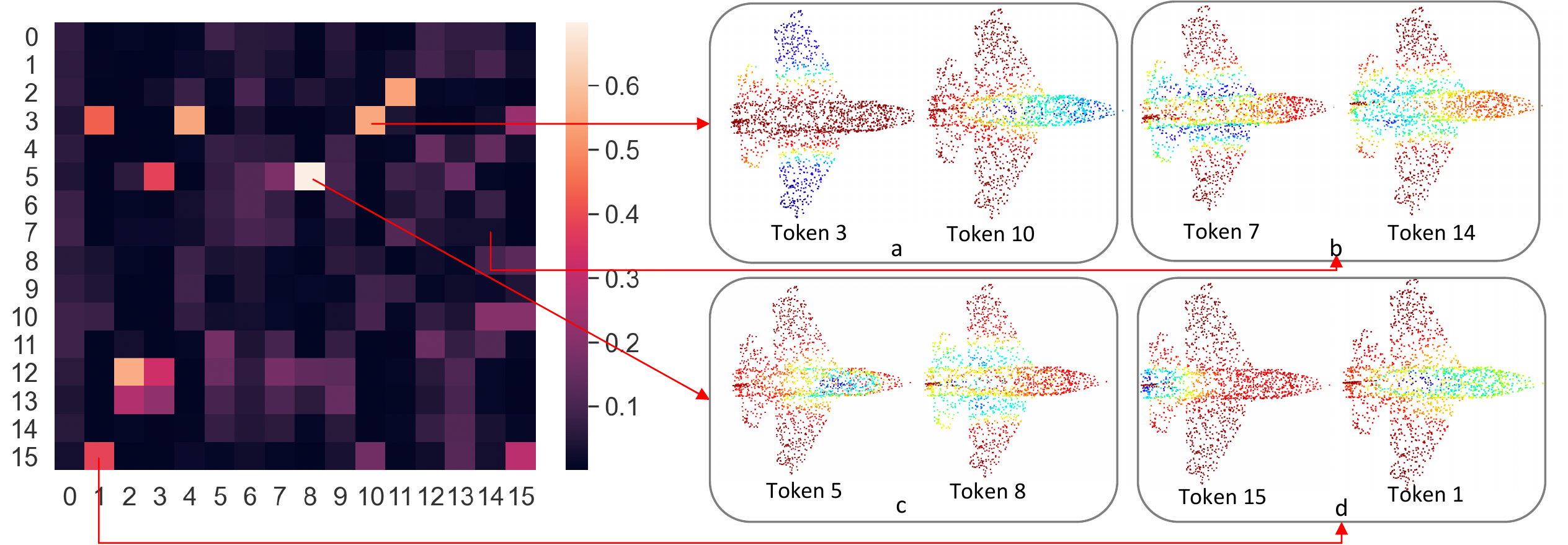}
    \caption{Visualization of the coefficient matrix (left) in self-attention module and the corresponding attentions to different parts (right). Blue (\textit{resp.} red) means the response is small (\textit{resp.} large). The figure (a), (b), (c), (d) show the token 3 (\textit{resp.} token 10) attending the body (\textit{resp.} the wings and tail), the token 7 (\textit{resp.} token 14) attending the wings and head (\textit{resp.} the wings and head), the token 5 (\textit{resp.} token 8) attending the wings and tail (\textit{resp.} the wings and head), the token 15 (\textit{resp.} token 1) and the wings (\textit{resp.} head to the wings and tail), respectively. We can observe that the token 3 and the token 10 in figure (a), and token 5 and token 8 in figure (c) correspond to large coefficient value since their attending parts have strong structure relation, which can form a airplane. Yet token 7 and token 14 in figure (b) correspond to small coefficient value since they attend very similar part thus have less structure relation.}
    \label{fig:cm}
\end{figure}

We also visualize the coefficient matrix in the self-attention module, for better visualization, we choose the models with $L$ equal to 16. As shown in Fig.~\ref{fig:cm}, the figure (a) in the right shows the token attending the body has more relations to the token indicating the flank and the wings, which together form a whole airplane. The figure (c) and (d) on the right are also similar, they attend different sub-regions which are connected together and build the structure of the airplane point-cloud. We find that for the tokens attending the relatively same parts, their coefficient values are low, which indicates that the self-attention module does not simply learn the similarity of different tokens instead capture the relations of different sub-regions that build the structure of a point-cloud. This can also be indicated from the small coefficient values between the same tokens.

\begin{table}[!t]
\centering
\small
\begin{tabular}{ c c   }
\hline
 Method &  Mean IoU \\
 \hline

 \textit{YOGO} w/o CA (concat global feature) & 84.1\\
 \textit{YOGO} w/o CA (adding tokens) & 84.7\\
 \textit{YOGO} w/o CA (concat tokens) & 84.6\\
 \textit{YOGO} w/o SA & 84.7 \\
 \hline
  \textit{YOGO} & 85.2\\

 \hline

\end{tabular}
\caption{Ablation study for self-attention module and cross-attention module in RIM. \textit{YOGO} represents that we are using the full modules, \textit{YOGO} w/o SA means no self-attention module, \textit{YOGO} w/o CA means no cross-attention. }
\label{tab:saca}
\end{table}

\textbf{Other techniques.} We study other techniques related to our proposed \textit{YOGO} including the effectiveness of choices of $L$ and $K$, the sampling and grouping methods, and the different pooling strategies.

\begin{figure}[!t]
    \centering
    \includegraphics[width=8.5cm]{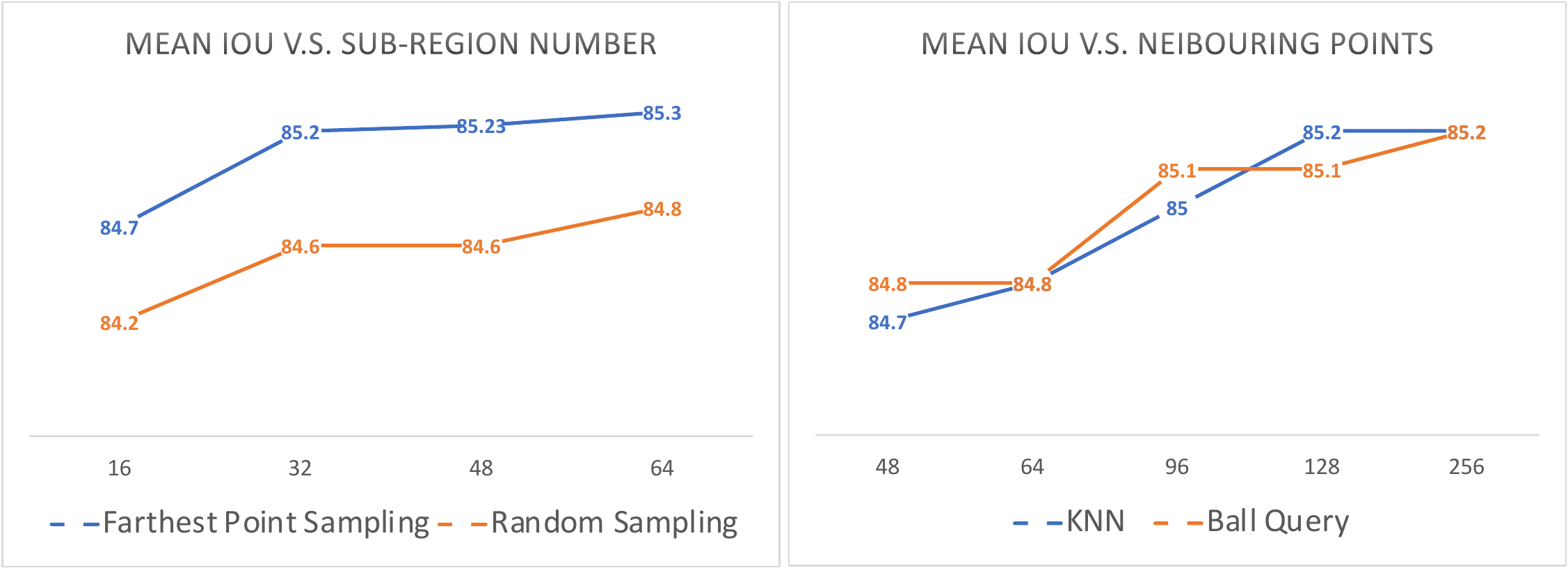}
    \caption{The left figure shows the effect of the number of sub-regions regarding the farthest point sampling and random sampling and the right figure shows the effect of the number of grouping points regarding the KNN grouping and ball query grouping.}
    \label{fig:miouvsKL}
\end{figure}

Regarding the choices of $L$ and $K$, we conduct experiments by fixing the $K$ equal to 64 and increasing the $L$ under both farthest point sampling (FPS) and random sampling, as well as fixing the $L$ equal to 32 and increasing the $K$ under both KNN and ball query \cite{qi2017pointnet++}, as presented in the Fig. \ref{fig:miouvsKL}. We find that when using different sampling methods, the performance will be better if increasing $L$. Besides, even for random sampling strategy, it has relatively small margin to farthest point sampling, which may indicate that our proposed \textit{YOGO} is robust to different sampling methods. When using different grouping method, the performance also consistently improves when $K$ become larger. Besides, we find there is not too much performance gap between ball query and KNN, this indicate our method is robust to different grouping methods.

Regarding the choices of different pooling operations, we study the differences between average pooling and max pooling, as shown in Table. \ref{tab:sf}. The experiment demonstrates the max pooling achieves higher performance than the average pooling. The reason is that the max pooling can better aggregate the most important information in a sub-region, which is more important for our proposed \textit{YOGO} since it highly depends on their relations.

% \begin{table}[!t]
% \centering
% \small
% \begin{tabular}{ c c c}
% \hline
% Part number &  FPS (mIoU) & Random sample (mIoU) \\
%  \hline
%  16 & 84.7 & 84.2  \\
%  32 & 85.2 & 84.6 \\
%  48 & 85.2 & 84.6  \\
%  64 & 85.3 & 84.8 \\
%  \hline
% \end{tabular}
% \caption{Ablation study for sampling the central points.}
% \label{tab:saca}
% \end{table}

\begin{table}[!t]
\centering
\small
\begin{tabular}{ c c c  }
\hline
 Method &  Mean IoU \\
 \hline
 \textit{YOGO} with max-pool& 85.2\\
 \textit{YOGO} with avg-pool& 85.0 \\

 \hline

\end{tabular}
\caption{Ablation study for different pooling strategies.}
\label{tab:sf}
\end{table}

\section{Discussion and conclusion}
In this paper, we propose a novel, simple, and elegant framework \textit{YOGO} (you only group once) for efficient point-cloud processing. Different from previous point-based methods, \textit{YOGO} divides point-cloud into a few sub-regions applying sampling and grouping operation once, leverages efficient token representation, models token relations and projects token to obtain strong point-wise features. \textit{YOGO} achieves at least 3x speedup over popular point-based baselines. Meanwhile, \textit{YOGO} delivers competitive performance on semantic segmentation on ShapeNetParts and S3DIS dataset, and classification on ModelNet40 dataset. Moreover, \textit{YOGO} is also robust to different sampling and grouping strategies, even for random sampling. It is noteworthy to mention that \textit{YOGO} introduced in this paper is a new framework and has huge potentials for point-cloud processing. In particular, it can be improved from the aspects of how to better sample then group out sub-regions, how to obtain better tokens, and how to better combine tokens and point-wise features etc. We hope that it can inspire the research community to further develop this framework.

\bibliographystyle{IEEEtran}

\bibliography{IEEEfull}

% Generated by IEEEtran.bst, version: 1.14 (2015/08/26)
\begin{thebibliography}{10}
\providecommand{\url}[1]{#1}
\csname url@samestyle\endcsname
\providecommand{\newblock}{\relax}
\providecommand{\bibinfo}[2]{#2}
\providecommand{\BIBentrySTDinterwordspacing}{\spaceskip=0pt\relax}
\providecommand{\BIBentryALTinterwordstretchfactor}{4}
\providecommand{\BIBentryALTinterwordspacing}{\spaceskip=\fontdimen2\font plus
\BIBentryALTinterwordstretchfactor\fontdimen3\font minus
  \fontdimen4\font\relax}
\providecommand{\BIBforeignlanguage}[2]{{%
\expandafter\ifx\csname l@#1\endcsname\relax
\typeout{** WARNING: IEEEtran.bst: No hyphenation pattern has been}%
\typeout{** loaded for the language `#1'. Using the pattern for}%
\typeout{** the default language instead.}%
\else
\language=\csname l@#1\endcsname
\fi
#2}}
\providecommand{\BIBdecl}{\relax}
\BIBdecl

\bibitem{qi2017pointnet}
C.~R. Qi, H.~Su, K.~Mo, and L.~J. Guibas, ``Pointnet: Deep learning on point
  sets for 3d classification and segmentation,'' in \emph{CVPR}, 2017.

\bibitem{qi2017pointnet++}
C.~R. Qi, L.~Yi, H.~Su, and L.~J. Guibas, ``Pointnet++: Deep hierarchical
  feature learning on point sets in a metric space,'' in \emph{NIPS}, 2017.

\bibitem{li2018pointcnn}
Y.~Li, R.~Bu, M.~Sun, W.~Wu, X.~Di, and B.~Chen, ``Pointcnn: Convolution on
  $\chi$-transformed points,'' in \emph{Proceedings of the 32nd International
  Conference on Neural Information Processing Systems}, 2018, pp. 828--838.

\bibitem{liu2019densepoint}
Y.~Liu, B.~Fan, G.~Meng, J.~Lu, S.~Xiang, and C.~Pan, ``Densepoint: Learning
  densely contextual representation for efficient point cloud processing,'' in
  \emph{Proceedings of the IEEE/CVF International Conference on Computer
  Vision}, 2019, pp. 5239--5248.

\bibitem{thomas2019KPConv}
H.~Thomas, C.~R. Qi, J.-E. Deschaud, B.~Marcotegui, F.~Goulette, and L.~J.
  Guibas, ``Kpconv: Flexible and deformable convolution for point clouds,''
  \emph{Proceedings of the IEEE International Conference on Computer Vision},
  2019.

\bibitem{wu2020visual}
B.~Wu, C.~Xu, X.~Dai, A.~Wan, P.~Zhang, M.~Tomizuka, K.~Keutzer, and P.~Vajda,
  ``Visual transformers: Token-based image representation and processing for
  computer vision,'' \emph{arXiv preprint arXiv:2006.03677}, 2020.

\bibitem{vaswani2017attention}
A.~Vaswani, N.~Shazeer, N.~Parmar, J.~Uszkoreit, L.~Jones, A.~N. Gomez,
  L.~Kaiser, and I.~Polosukhin, ``Attention is all you need,'' 2017.

\bibitem{maturana2015voxnet}
D.~Maturana and S.~Scherer, ``Voxnet: A 3d convolutional neural network for
  real-time object recognition,'' in \emph{2015 IEEE/RSJ International
  Conference on Intelligent Robots and Systems (IROS)}.\hskip 1em plus 0.5em
  minus 0.4em\relax IEEE, 2015, pp. 922--928.

\bibitem{ben20183dmfv}
Y.~Ben-Shabat, M.~Lindenbaum, and A.~Fischer, ``3dmfv: Three-dimensional point
  cloud classification in real-time using convolutional neural networks,''
  \emph{IEEE Robotics and Automation Letters}, vol.~3, no.~4, pp. 3145--3152,
  2018.

\bibitem{roynard2018classification}
X.~Roynard, J.-E. Deschaud, and F.~Goulette, ``Classification of point cloud
  scenes with multiscale voxel deep network,'' \emph{arXiv preprint
  arXiv:1804.03583}, 2018.

\bibitem{liu2019point}
Z.~Liu, H.~Tang, Y.~Lin, and S.~Han, ``Point-voxel cnn for efficient 3d deep
  learning,'' \emph{arXiv preprint arXiv:1907.03739}, 2019.

\bibitem{su2018splatnet}
H.~Su, V.~Jampani, D.~Sun, S.~Maji, E.~Kalogerakis, M.-H. Yang, and J.~Kautz,
  ``Splatnet: Sparse lattice networks for point cloud processing,'' in
  \emph{Proceedings of the IEEE conference on computer vision and pattern
  recognition}, 2018, pp. 2530--2539.

\bibitem{liu2015sparse}
B.~Liu, M.~Wang, H.~Foroosh, M.~Tappen, and M.~Pensky, ``Sparse convolutional
  neural networks,'' in \emph{Proceedings of the IEEE conference on computer
  vision and pattern recognition}, 2015, pp. 806--814.

\bibitem{choy20194d}
C.~Choy, J.~Gwak, and S.~Savarese, ``4d spatio-temporal convnets: Minkowski
  convolutional neural networks,'' in \emph{Proceedings of the IEEE/CVF
  Conference on Computer Vision and Pattern Recognition}, 2019, pp. 3075--3084.

\bibitem{tang2020searching}
H.~Tang, Z.~Liu, S.~Zhao, Y.~Lin, J.~Lin, H.~Wang, and S.~Han, ``Searching
  efficient 3d architectures with sparse point-voxel convolution,'' in
  \emph{European Conference on Computer Vision}.\hskip 1em plus 0.5em minus
  0.4em\relax Springer, 2020, pp. 685--702.

\bibitem{zhou2020cylinder3d}
H.~Zhou, X.~Zhu, X.~Song, Y.~Ma, Z.~Wang, H.~Li, and D.~Lin, ``Cylinder3d: An
  effective 3d framework for driving-scene lidar semantic segmentation,''
  \emph{arXiv preprint arXiv:2008.01550}, 2020.

\bibitem{yan2018second}
Y.~Yan, Y.~Mao, and B.~Li, ``Second: Sparsely embedded convolutional
  detection,'' \emph{Sensors}, vol.~18, no.~10, p. 3337, 2018.

\bibitem{wang2018fusing}
Z.~Wang, W.~Zhan, and M.~Tomizuka, ``Fusing bird’s eye view lidar point cloud
  and front view camera image for 3d object detection,'' in \emph{2018 IEEE
  Intelligent Vehicles Symposium (IV)}.\hskip 1em plus 0.5em minus 0.4em\relax
  IEEE, 2018, pp. 1--6.

\bibitem{wu2017squeezeseg}
B.~Wu, A.~Wan, X.~Yue, and K.~Keutzer, ``Squeezeseg: Convolutional neural nets
  with recurrent crf for real-time road-object segmentation from 3d lidar point
  cloud,'' in \emph{ICRA}, 2018.

\bibitem{wu2018squeezesegv2}
B.~Wu, X.~Zhou, S.~Zhao, X.~Yue, and K.~Keutzer, ``Squeezesegv2: Improved model
  structure and unsupervised domain adaptation for road-object segmentation
  from a lidar point cloud,'' in \emph{ICRA}, 2019.

\bibitem{xu2020squeezesegv3}
C.~Xu, B.~Wu, Z.~Wang, W.~Zhan, P.~Vajda, K.~Keutzer, and M.~Tomizuka,
  ``Squeezesegv3: Spatially-adaptive convolution for efficient point-cloud
  segmentation,'' in \emph{European Conference on Computer Vision}.\hskip 1em
  plus 0.5em minus 0.4em\relax Springer, 2020, pp. 1--19.

\bibitem{su2015multi}
H.~Su, S.~Maji, E.~Kalogerakis, and E.~Learned-Miller, ``Multi-view
  convolutional neural networks for 3d shape recognition,'' in
  \emph{Proceedings of the IEEE international conference on computer vision},
  2015, pp. 945--953.

\bibitem{lawin2017deep}
F.~J. Lawin, M.~Danelljan, P.~Tosteberg, G.~Bhat, F.~S. Khan, and M.~Felsberg,
  ``Deep projective 3d semantic segmentation,'' in \emph{International
  Conference on Computer Analysis of Images and Patterns}.\hskip 1em plus 0.5em
  minus 0.4em\relax Springer, 2017, pp. 95--107.

\bibitem{boulch2017unstructured}
A.~Boulch, B.~Le~Saux, and N.~Audebert, ``Unstructured point cloud semantic
  labeling using deep segmentation networks.'' \emph{3DOR}, vol.~2, p.~7, 2017.

\bibitem{yang2018pixor}
B.~Yang, W.~Luo, and R.~Urtasun, ``Pixor: Real-time 3d object detection from
  point clouds,'' in \emph{Proceedings of the IEEE conference on Computer
  Vision and Pattern Recognition}, 2018, pp. 7652--7660.

\bibitem{lang2019pointpillars}
A.~H. Lang, S.~Vora, H.~Caesar, L.~Zhou, J.~Yang, and O.~Beijbom,
  ``Pointpillars: Fast encoders for object detection from point clouds,'' in
  \emph{Proceedings of the IEEE/CVF Conference on Computer Vision and Pattern
  Recognition}, 2019, pp. 12\,697--12\,705.

\bibitem{milioto2019rangenet++}
A.~Milioto, I.~Vizzo, J.~Behley, and C.~Stachniss, ``Rangenet++: Fast and
  accurate lidar semantic segmentation,'' in \emph{2019 IEEE/RSJ International
  Conference on Intelligent Robots and Systems (IROS)}.\hskip 1em plus 0.5em
  minus 0.4em\relax IEEE, 2019, pp. 4213--4220.

\bibitem{armeni2017joint}
I.~Armeni, S.~Sax, A.~R. Zamir, and S.~Savarese, ``Joint 2d-3d-semantic data
  for indoor scene understanding,'' \emph{arXiv preprint arXiv:1702.01105},
  2017.

\bibitem{chang2015shapenet}
A.~X. Chang, T.~Funkhouser, L.~Guibas, P.~Hanrahan, Q.~Huang, Z.~Li,
  S.~Savarese, M.~Savva, S.~Song, H.~Su \emph{et~al.}, ``Shapenet: An
  information-rich 3d model repository,'' \emph{arXiv preprint
  arXiv:1512.03012}, 2015.

\bibitem{hua2018pointwise}
B.-S. Hua, M.-K. Tran, and S.-K. Yeung, ``Pointwise convolutional neural
  networks,'' in \emph{Proceedings of the IEEE Conference on Computer Vision
  and Pattern Recognition}, 2018, pp. 984--993.

\bibitem{liu2020closer}
Z.~Liu, H.~Hu, Y.~Cao, Z.~Zhang, and X.~Tong, ``A closer look at local
  aggregation operators in point cloud analysis,'' in \emph{European Conference
  on Computer Vision}.\hskip 1em plus 0.5em minus 0.4em\relax Springer, 2020,
  pp. 326--342.

\bibitem{wang2017cnn}
P.-S. Wang, Y.~Liu, Y.-X. Guo, C.-Y. Sun, and X.~Tong, ``O-cnn: Octree-based
  convolutional neural networks for 3d shape analysis,'' \emph{ACM Transactions
  on Graphics (TOG)}, vol.~36, no.~4, pp. 1--11, 2017.

\bibitem{li2018so}
J.~Li, B.~M. Chen, and G.~H. Lee, ``So-net: Self-organizing network for point
  cloud analysis,'' in \emph{Proceedings of the IEEE conference on computer
  vision and pattern recognition}, 2018, pp. 9397--9406.

\bibitem{komarichev2019cnn}
A.~Komarichev, Z.~Zhong, and J.~Hua, ``A-cnn: Annularly convolutional neural
  networks on point clouds,'' in \emph{Proceedings of the IEEE/CVF Conference
  on Computer Vision and Pattern Recognition}, 2019, pp. 7421--7430.

\bibitem{carion2020end}
N.~Carion, F.~Massa, G.~Synnaeve, N.~Usunier, A.~Kirillov, and S.~Zagoruyko,
  ``End-to-end object detection with transformers,'' in \emph{European
  Conference on Computer Vision}.\hskip 1em plus 0.5em minus 0.4em\relax
  Springer, 2020, pp. 213--229.

\bibitem{dosovitskiy2020image}
A.~Dosovitskiy, L.~Beyer, A.~Kolesnikov, D.~Weissenborn, X.~Zhai,
  T.~Unterthiner, M.~Dehghani, M.~Minderer, G.~Heigold, S.~Gelly \emph{et~al.},
  ``An image is worth 16x16 words: Transformers for image recognition at
  scale,'' \emph{arXiv preprint arXiv:2010.11929}, 2020.

\bibitem{peize2020sparse}
P.~Sun, R.~Zhang, Y.~Jiang, T.~Kong, C.~Xu, W.~Zhan, M.~Tomizuka, L.~Li,
  Z.~Yuan, C.~Wang, and P.~Luo, ``{SparseR-CNN}: End-to-end object detection
  with learnable proposals,'' \emph{arXiv preprint arXiv:2011.12450}, 2020.

\bibitem{zhu2019asymmetric}
Z.~Zhu, M.~Xu, S.~Bai, T.~Huang, and X.~Bai, ``Asymmetric non-local neural
  networks for semantic segmentation,'' in \emph{Proceedings of the IEEE/CVF
  International Conference on Computer Vision}, 2019, pp. 593--602.

\bibitem{devlin2018bert}
J.~Devlin, M.-W. Chang, K.~Lee, and K.~Toutanova, ``Bert: Pre-training of deep
  bidirectional transformers for language understanding,'' \emph{arXiv preprint
  arXiv:1810.04805}, 2018.

\bibitem{shilane2004princeton}
P.~Shilane, P.~Min, M.~Kazhdan, and T.~Funkhouser, ``The princeton shape
  benchmark,'' in \emph{Proceedings Shape Modeling Applications, 2004.}\hskip
  1em plus 0.5em minus 0.4em\relax IEEE, 2004, pp. 167--178.

\bibitem{yanx}
Y.~Xu, ``Pointnet\_pointnet2\_pytorch,''
  \url{https://github.com/yanx27/Pointnet_Pointnet2_pytorch}, 2019.

\bibitem{huang2018recurrent}
Q.~Huang, W.~Wang, and U.~Neumann, ``Recurrent slice networks for 3d
  segmentation of point clouds,'' in \emph{Proceedings of the IEEE Conference
  on Computer Vision and Pattern Recognition}, 2018, pp. 2626--2635.

\bibitem{yi2017syncspeccnn}
L.~Yi, H.~Su, X.~Guo, and L.~J. Guibas, ``Syncspeccnn: Synchronized spectral
  cnn for 3d shape segmentation,'' in \emph{Proceedings of the IEEE Conference
  on Computer Vision and Pattern Recognition}, 2017, pp. 2282--2290.

\bibitem{wang2019dynamic}
Y.~Wang, Y.~Sun, Z.~Liu, S.~E. Sarma, M.~M. Bronstein, and J.~M. Solomon,
  ``Dynamic graph cnn for learning on point clouds,'' \emph{Acm Transactions On
  Graphics (tog)}, vol.~38, no.~5, pp. 1--12, 2019.

\bibitem{xu2018spidercnn}
Y.~Xu, T.~Fan, M.~Xu, L.~Zeng, and Y.~Qiao, ``Spidercnn: Deep learning on point
  sets with parameterized convolutional filters,'' in \emph{Proceedings of the
  European Conference on Computer Vision (ECCV)}, 2018, pp. 87--102.

\bibitem{tatarchenko2018tangent}
M.~Tatarchenko, J.~Park, V.~Koltun, and Q.-Y. Zhou, ``Tangent convolutions for
  dense prediction in 3d,'' in \emph{Proceedings of the IEEE Conference on
  Computer Vision and Pattern Recognition}, 2018, pp. 3887--3896.

\bibitem{hermosilla2018monte}
P.~Hermosilla, T.~Ritschel, P.-P. V{\'a}zquez, {\`A}.~Vinacua, and T.~Ropinski,
  ``Monte carlo convolution for learning on non-uniformly sampled point
  clouds,'' \emph{ACM Transactions on Graphics (TOG)}, vol.~37, no.~6, pp.
  1--12, 2018.

\bibitem{jaegle2021perceiver}
A.~Jaegle, F.~Gimeno, A.~Brock, A.~Zisserman, O.~Vinyals, and J.~Carreira,
  ``Perceiver: General perception with iterative attention,'' 2021.

\end{thebibliography}

\end{document}